\pdfoutput=1

\documentclass[11pt]{article}

\usepackage{ACL2023}

\usepackage{times}
\usepackage{latexsym}

\usepackage[T1]{fontenc}

\usepackage[utf8]{inputenc}

\usepackage{microtype}

\usepackage{inconsolata}

\usepackage{booktabs}
\usepackage{comment}
\usepackage{multirow}
\usepackage{graphicx}
\usepackage{bm}
\usepackage{amsmath}

\makeatletter
\newcommand\footnoteref[1]{\protected@xdef\@thefnmark{\ref{#1}}\@footnotemark}
\makeatother

\newcommand{\yn}[1]{\textcolor{black}{#1}}

\title{A Unified Generative Approach to Product Attribute-Value Identification}

\author{Keiji Shinzato\\
Rakuten Institute of Technology,\\
Rakuten Group, Inc.\\
\texttt{keiji.shinzato@rakuten.com}
\And Naoki Yoshinaga\\
Institute of Industrial Science,\\
The University of Tokyo\\
\texttt{ynaga@iis.u-tokyo.ac.jp}
\AND Yandi Xia\\
Rakuten Institute of Technology,\\
Rakuten Group, Inc.\\
\texttt{yandi.xia@rakuten.com}
\And Wei-Te Chen\\
Rakuten Institute of Technology,\\
Rakuten Group, Inc.\\
\texttt{weite.chen@rakuten.com}}

\begin{document}
\maketitle

\begin{abstract}
Product attribute-value identification (\textsc{pavi}) has been studied to link  
products on e-commerce sites with their attribute values (\textit{e.g.}, $\langle$Material,
Cotton$\rangle$) using product text as clues.
Technical demands from real-world e-commerce platforms require
\textsc{pavi} methods to handle unseen values,
multi-attribute values, and canonicalized values, which are only partly
addressed in existing extraction- and classification-based
approaches. 
Motivated by this, we explore a
generative approach to the \textsc{pavi} task. We
finetune a
pre-trained generative model, \textsc{t5}, 
to decode a set of attribute-value pairs
as a target sequence from the given product text.  Since the
attribute-value pairs are unordered set elements, how to linearize them will matter; we, thus, explore methods of composing an attribute-value pair and ordering the pairs for the task.
Experimental results confirm that our generation-based
approach outperforms the existing extraction- and classification-based methods on large-scale real-world datasets meant for those methods.
\end{abstract}

\section{Introduction}
\begin{figure}
    \centering
    \includegraphics[width=\columnwidth]{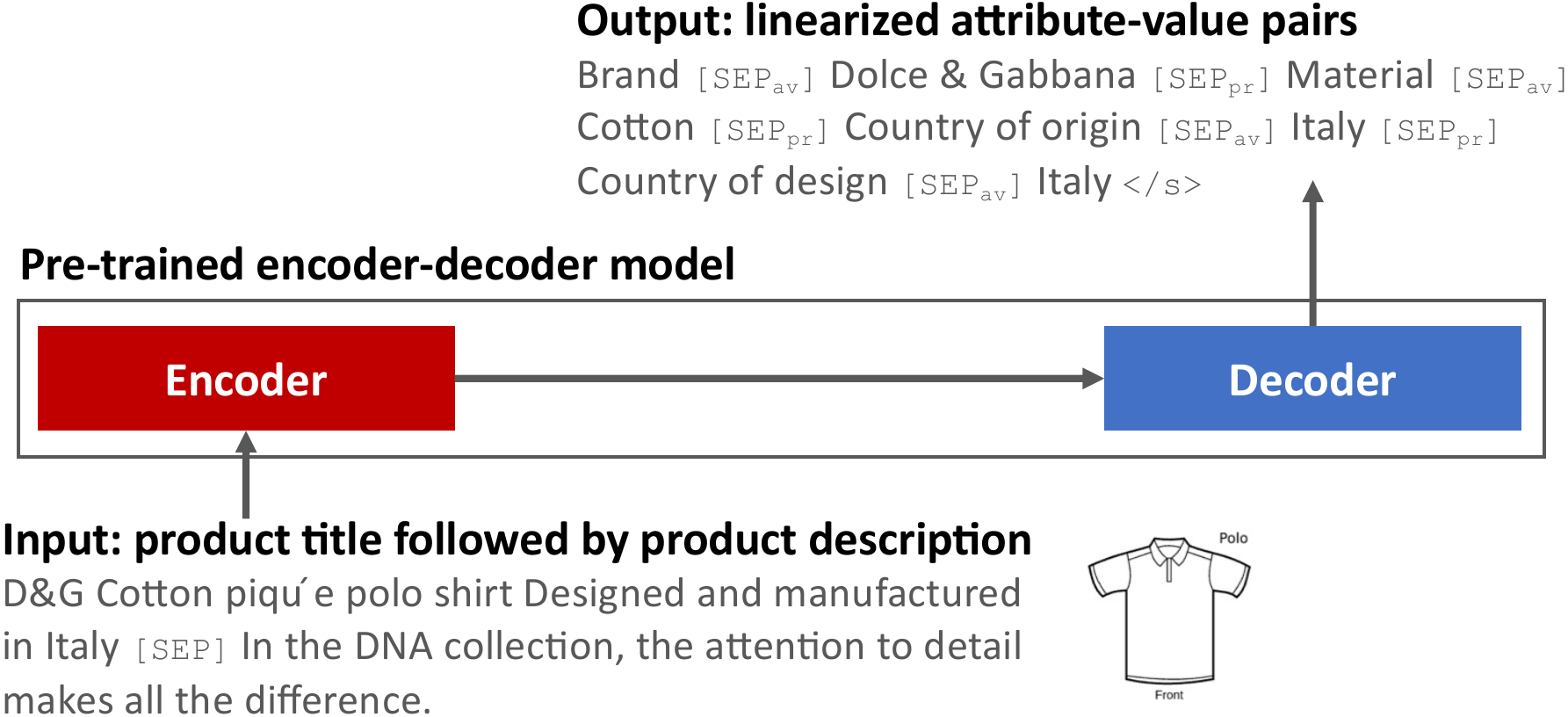}
    \caption{Overview of our 
    generative approach for \textsc{pavi}; 
    it takes product text
    to return a set of attribute-value pairs. In this example, the model generates \textit{Dolce \& Gabbana} as a brand, which is a canonicalized form of \textit{D\&G}, and two attributes
    have the entity \textit{Italy} as values.}
    \label{fig:overview}
\end{figure}

Since organized product data play a crucial role in serving better product search and recommendation to customers, product attribute value identification (\textsc{pavi}) has been a core task in the e-commerce industry.
For attributes pre-defined by e-commerce sites, the task aims to link values of those attributes to products using 
product titles and descriptions as clues (Figure~\ref{fig:overview}). For example, 
from the title ``\textit{D\&G Cotton piqué polo shirt Designed and manufactured in Italy},'' models are required to return a set of possible attribute-value pairs, namely $\lbrace\langle$Brand, \textit{Dolce \& Gabbana}$\rangle$, $\langle$Material, \textit{Cotton}$\rangle$, $\langle$Country of origin, \textit{Italy}$\rangle$, $\langle$Country of design, \textit{Italy}$\rangle\rbrace$.

In the literature,
\textsc{pavi} has been addressed basically by extraction from the product text by using named entity recognition~\citep{probst_2007,wong_2008,putthividhya_2011,bing_2012,shinzato_2013,more_2016,zheng_2018,rezk_2019,karamanolakis_2020,zhang_2020}
or question answering~\citep{xu_2019,wang_2020,shinzato_2022,mave}.
However, since \textsc{pavi} requires canonicalized values rather than raw value strings in the product text, 
some researchers have started to solve \textsc{pavi} as classification~\citep{chen_2022,fuchs_2022}.

%
To adopt \textsc{pavi} models in real-world e-commerce platforms, there are the following challenges.

\paragraph{Unseen values.} Since values can be entities
such as \textit{brands},
models need to identify values unseen in the training data~\citep{zheng_2018}.
Since the classification-based approach assumes a pre-defined set of target classes (attribute-value pairs), it cannot
handle such unseen attribute-value pairs.

\paragraph{Multi-attribute
values.} When values can be associated with multiple attributes (\textit{e.g.}, \textit{Italy} in Figure~\ref{fig:overview}), 
models need to identify multiple attributes for a single value string in the text.
%
To address this, the extraction-based approach must solve nested named entity recognition~\citep{wang_2020}.

\paragraph{Canonicalized values.} 
E-commerce vendors need attribute values in the canonical form (\textit{e.g., Dolce \& Gabbana} for \textit{D\&G}) in actual services such as faceted product search~\cite{chen_2022}.
The extraction-based approach needs a further step to canonicalize extracted raw value strings~\citep{putthividhya_2011,zhang_2021}.

\begin{table}[t]
    \centering
    \small
    \begin{tabular}{lccc}
    \toprule
    Approach & Unseen & Multi & Canon\\
    \midrule
    Extraction     & Support & Partially & Not\\
    Classification & Not & Support & Support\\
    Generation (ours)     & Support & Support & Support\\
    \bottomrule
    \end{tabular}
    \caption{Comparison of different \textsc{pavi} approaches in terms of the challenges: handling \textbf{unseen} values, \textbf{multi}-attribute values, and \textbf{canon}icalized values.}
    \label{tab:summary_of_requirements}
\end{table}

\noindent

Motivated by 
the shortcomings of 
the existing approaches 
to \textsc{pavi} (Table~\ref{tab:summary_of_requirements}),
we propose 
to cast 
\textsc{pavi} 
as sequence-to-set generation, which
can handle all the challenges by using canonicalized attribute-value pairs for training (Figure~\ref{fig:overview}).
We expect that 1) generation can decode unseen values by considering corresponding values in the input, 2) generation can decode the same string in the input multiple times as values for different attributes, and 3) generation can learn how to canonicalize raw strings in input.
%
We finetune the pre-trained 
generative model \textsc{t5}~\citep{t5}
to autoregressively decode a set of attribute-value pairs from the given text. 
%
As discussed in \citep{vinjal2016,yang_2018,madaan_2022}, the output order will matter to 
decode sets as a sequence.
We therefore explore methods of composing an attribute-value pair and ordering the pairs 
for the task.

We 
evaluate
our 
generative framework 
on two real-world datasets, 
\textsc{mave}~\citep{mave} and our in-house product data.
The experimental results demonstrate that our generation-based approach
outperforms 
extraction- and classification-based methods on their target datasets.

Our contribution is as follows.

\begin{itemize}
    \item We have solved the product attribute-value identification task as a sequence-to-set generation for the first time. 
    \item 
    We revealed the effective
    order of attribute-value pairs for the \textsc{t5} model among various ordering schemes (Table~\ref{tab:example_of_av_seq_order}).
    \item We provided the first comprehensive comparison among extraction-, classification-, and generation-based models on two real-world \textsc{pavi}  datasets, and empirically confirmed that the generation-based models outperformed the others
    (Table~\ref{tab:old_performance})
while addressing  all challenges in \textsc{pavi} (Tables~\ref{tab:unseen_performance}, \ref{tab:performance_on_nested_values} and \ref{tab:performance_on_canonicalized_values}).
\end{itemize}

\section{Related Work}
\paragraph{Product Attribute-Value Extraction}
Traditionally, a myriad of previous studies formulated \textsc{pavi} as named entity recognition (\textsc{ner})~\citep{probst_2007,wong_2008,putthividhya_2011,bing_2012,shinzato_2013,more_2016,zheng_2018,rezk_2019,karamanolakis_2020,zhang_2020}. However, since the number of attributes in real-world e-commerce sites can exceed ten thousand~\citep{xu_2019}, the \textsc{ner}-based 
models suffer from the data sparseness problem, which makes the models perform poorly. 
While the extraction-based approach can identify unseen values in the training data, it cannot canonicalize values by itself and is difficult to handle overlapping values, although nested \textsc{ner} (surveyed in~\citet{10.1145/3522593}) can 
remedy the latter issue. 

\begin{table*}[t]
\small
    \centering
    \begin{tabular}{ll}\toprule
    Ordering & Attribute-value pairs placed in the target sequence\\\midrule
    Rare-first  &
    \textcolor{black}{\textrm{Material}} \textsc{[sep$_{\mathit{av}}$]} \textcolor{black}{Nylon} \textsc{[sep$_{\mathit{pr}}$]}
    \textcolor{black}{\textrm{Color}} \textsc{[sep$_{\mathit{av}}$]} \textcolor{black}{Red} \textsc{[sep$_{\mathit{pr}}$]}
    \textcolor{black}{\textrm{Color}} \textsc{[sep$_{\mathit{av}}$]} \textcolor{black}{White}\\

    Common-first &
    \textcolor{black}{\textrm{Color}} \textsc{[sep$_{\mathit{av}}$]} \textcolor{black}{White} \textsc{[sep$_{\mathit{pr}}$]}
    \textcolor{black}{\textrm{Color}} \textsc{[sep$_{\mathit{av}}$]} \textcolor{black}{Red} \textsc{[sep$_{\mathit{pr}}$]}
    \textcolor{black}{\textrm{Material}} \textsc{[sep$_{\mathit{av}}$]} \textcolor{black}{Nylon}\\
    Random&
    \textcolor{black}{\textrm{Color}} \textsc{[sep$_{\mathit{av}}$]} \textcolor{black}{Red} \textsc{[sep$_{\mathit{pr}}$]}
    \textcolor{black}{\textrm{Material}} \textsc{[sep$_{\mathit{av}}$]} \textcolor{black}{Nylon} \textsc{[sep$_{\mathit{pr}}$]}
    \textcolor{black}{\textrm{Color}} \textsc{[sep$_{\mathit{av}}$]} \textcolor{black}{White}\\
    
    \bottomrule    
    \end{tabular}
    \caption{Example of attribute-value pair ordering with the attribute-then-value composition.
    We assume that the frequency of the pairs is $\langle$Color, White$\rangle > \langle$Color, Red$\rangle > \langle$Material, Nylon$\rangle$. 
}
    \label{tab:example_of_av_seq_order}
\end{table*}

To mitigate the data sparseness problem, some studies leveraged \textsc{qa} models for the \textsc{pavi} task~\citep{xu_2019,wang_2020,mave,shinzato_2022}, by assuming the target attribute for extraction as additional input. These \textsc{qa}-based approaches take an attribute as \textit{query} and product text as \textit{context}, and extract attribute values from the context as \textit{answer} for the query. 
%
%
Similar to the traditional \textsc{ner}-based models, these extractive \textsc{qa}-based models do not work for canonicalized values.
To improve the ability to find unseen values, \citet{roy_2021} generated a value for the given product text and attribute.
However, 
we need to apply these \textsc{qa}-based models to the same context with each of thousands of attributes, unless comprehensive attribute taxonomy is designed to narrow down possible attributes; such taxonomy is not always available and is often imperfect, as investigated by~\citet{mao_2020} for Amazon.com.

%

\paragraph{Product Attribute-Value Identification as Classification}
\citet{chen_2022} 
solved \textsc{pavi} as 
multi-label classification (\textsc{mlc}), assuming attribute-value pairs as target labels. One of the problems in this approach is that the distribution between positive and negative labels is heavily skewed because the number of possible attribute values per product is much smaller than the total number of attribute values. 
To alleviate the imbalanced label problem, they introduced a method called label masking to reduce the number of negative labels using an attribute taxonomy designed by the e-commerce platform. To mitigate the extreme multi-class classification, \citet{fuchs_2022} decomposed the target label, namely attribute-value pair, into two atomic labels, attribute and value, to perform a hierarchical classification. 
%
Although these classification-based approaches support canonicalized values and multi-attribute values, they cannot 
handle unseen values.



In this study, we adopt a generative approach to return a set of attribute-value pairs from given product data, and 
empirically compare it with the above two
approaches.
Our approach can be applied to the task settings adopted by the \textsc{qa}-based models, by simply feeding one (or more) target attributes as additional input (\textit{e.g.}, title [\textsc{sep}] description [\textsc{sep}] attributes) 
to decode their values in order.

\section{Proposed Method}

As mentioned above, previous studies formalize \textsc{pavi} as either sequence tagging or multi-label classification problems. These approaches do not address all the challenges derived from real-world e-commerce sites at the same time (Table~\ref{tab:summary_of_requirements}).

We thus propose a unified generative framework that formalizes \textsc{pavi} as a sequence-to-set problem.
Let us denote $\bm{x} = \{x_1,x_2,\ldots,x_n\}$ as product data (title and description) where $n$ is the number of tokens in $\bm{x}$.
%
Given product data $\bm{x}$, the model is trained to return a set of 
attribute-value pairs $\bm{y} = \{\langle\bm{a_1},\bm{v_1}\rangle, \langle\bm{a_2},\bm{v_2}\rangle, \dots, \langle\bm{a_k},\bm{v_k}\rangle\}$
for $\bm{x}$, where $k$ is the number of attribute-value pairs associated with the product;
$\bm{a}_i = \{a_{1},a_{2},\ldots,a_{m_i}\}$ and $\bm{v}_i = \{v_{1},v_{2},\ldots,v_{l_i}\}$ are corresponding attribute and value.\footnote{When there are more than one value for the same attribute (\textit{e.g.}, 
size), we decompose a pair of attribute and $n (> 1)$ multiple values into $n$ attribute-value pairs (Table~\ref{tab:example_of_product_data}).} $m_i$ and $l_i$ are the numbers of tokens in $\bm{a}_{i}$ and $\bm{v}_i$, respectively.

As the backbone of our 
approach, we employ \textsc{t5}~\citep{t5}, a pre-trained generative model based on Transformer~\citep{vaswani_2017} that maps an input sequence to an output sequence.
%

The key issue in formulating the \textsc{pavi} task as sequence-to-sequence generation is how to linearize a set of attribute-value pairs into a sequence.
Firstly, we should consider how to associate attributes and their corresponding values in the output sequence. Secondly, the autoregressive generation decodes output tokens (here, attributes and values) one by one conditioned on the previous labels. Thus, if specific (or informative) tokens are first decoded, it will make it easy to decode the remaining tokens. However, due to the exposure bias, decoding specific (namely, infrequent) tokens are more likely to fail.
To address the challenge, we decompose the issue on linearization 
into two subproblems on how to compose an attribute-value pair and how to order attribute-value pairs. 
In what follows, we will describe these subproblems.

\begin{table*}[t]
\small
    \centering
    \tabcolsep3.9pt
    \begin{tabular}{lrrrrrr}\toprule
    & \multicolumn{3}{c}{\textsc{mave}}
    & \multicolumn{3}{c}{In-House Product Data}\\
    \cmidrule(l{2pt}r{2pt}){2-4}
    \cmidrule(l{2pt}r{2pt}){5-7}
    & Train & Dev. & Test
    & Train & Dev. & Test\\\midrule
The number of examples &
640,000 & 100,000 & 290,773 &
640,000 & 100,000 & 100,000 \\
\qquad without values &
150,412 & 23,220 & 67,936 &
      0 &      0 &      0 \\
The number of distinct attributes &
693   & 660   & 685   &
1,320 & 1,119 & 1,123 \\
The number of distinct attribute values &
54,200 & 21,734 & 37,092 &
13,328 &  8,402 &  8,445 \\
The number of distinct attribute-value pairs &
63,715 & 25,675 & 43,605 &
14,829 &  9,310 &  9,356 \\
The number of attribute-value pairs &
1,594,855 & 249,543 & 722,130 &
2,966,227 & 463,463 & 462,507 \\
\qquad with unseen values &
0 & 4,667 & 13,578 &
0 &   443 &    491 \\
\qquad with multi-attribute (or nested) values &
134,290 & 20,832 & 60,832 &
103,727 & 16,280 & 15,843 \\
\qquad whose values appear as raw strings in the product text & 
1,594,855 & 249,543 & 722,130 &
1,340,043 & 210,181 & 207,997 \\
The average number of subwords per example (input) &
 253.73 & 253.73 & 253.56  &
 357.87 & 359.89 & 356.93 \\
The average number of subwords per example (output) &
  10.35 & 10.39 & 10.32 &
  46.43 & 46.44 & 46.31 \\
The average number of attributes per example &
1.64 & 1.64 & 1.64 & 
3.24 & 3.25 & 3.23 \\
The average number of values per example &
2.25 & 2.25 & 2.25 &
4.62 & 4.62 & 4.61 \\
The average number of subwords per attribute &
 2.84 & 2.82 & 2.85 &
 4.77 & 4.72 & 4.69 \\
The average number of subwords per value &
 4.15 & 3.46 & 3.81 &
 4.09 & 3.96 & 3.93 \\

\bottomrule
    \end{tabular}
    \caption{Detailed statistics of the datasets. We used the \textsc{t5} tokenizer to tokenize examples, attributes and values.}
    \label{tab:dataset_stats}
\end{table*}

\subsection{Composition of Attribute-Value Pair}
We consider the following ways to compose an attribute-value pair.\footnote{We have also attempted to generate all attributes prior to values 
(namely,  $\bm{a}_1[\textsc{sep}_{\mathit{pr}}]\ldots\bm{a}_k[\textsc{sep}_{\mathit{av}}]\bm{v}_1[\textsc{sep}_{\mathit{pr}}]\ldots\bm{v}_k$) 
or vice versa;
this unpaired generation slightly underperformed the paired generation used here.}
In both ways, attributes and values are separated by a special token \textsc{[sep$_{\mathit{av}}$]}.

\paragraph{Attribute-then-value, $\langle$A, V$\rangle$} Attribute is placed, and then its value (\textit{e.g.,} Color \textsc{[sep$_{\mathit{av}}$]} White). 
In general, the vocabulary size of attributes is much smaller than that of values. Thus, models will be easier to decode attributes than values.

\paragraph{Value-then-attribute, $\langle$V, A$\rangle$} Value is placed, and then its attribute (\textit{e.g.,} White \textsc{[sep$_{\mathit{av}}$]} Color). This will be effective when the target values appear as raw strings in the given text and are easier to decode than attributes.

\subsection{Ordering of Attribute-Value Pairs}
\label{av_pair_ordering}
In this work, we design three different types of the attribute-value pair ordering (Table~\ref{tab:example_of_av_seq_order}). We use a special token \textsc{[sep$_{\mathit{pr}}$]} as a separator between pairs.
\paragraph{Rare-first} 
Specific attribute values
(\textit{e.g.}, brands) can 
help models decode other attribute values.
For example, since \textit{Levi's} has many products made of denim, it is easy to 
decode the material if \textit{Levi's} is decoded 
in advance. Meanwhile, since there are many brands that have products made of denim, decoding denim as a material in advance is
useless
to decode the brands. To capture this
inter-value
dependency, we assume 
a correlation between the frequency and specificity of attribute-value pairs, and place attribute-value pairs to the target sequence in rare-first ordering of attribute-value pair frequency calculated from the training data. 
The attribute-value pairs with the same ranking will be placed randomly for this and following ordering.

\paragraph{Common-first} 
When the model autoregressively decodes outputs, intermediate errors affect future decoding.
Thus, it is important to decode from confident attribute-value pairs.
Since models 
will be easier
to decode 
attribute-value pairs that have more training examples,
we place attribute-value pairs to the target sequence in the common-first ordering of attribute-value pair frequency. This approach is adopted by~\citet{yang_2018} in solving multi-label document classification as generation.

\paragraph{Random}
To see whether the orders matter, we randomly sort attribute-value pairs in the target sequence; more precisely, 
we collect, uniquify, and shuffle attribute-value pairs taken from all training examples, and sort the pairs in each example according to the obtained order of the pairs.
If this random ordering shows inferior performance against the above orderings, we can conclude output orders matter in this task.

\section{Experiments}
We evaluate our generative approach to \textsc{pavi} using two real-world datasets.
In the literature, different types of approaches are rarely compared 
due to the proprietary nature of codes and datasets in this task. We thus compare our generation-based model with extraction- and classification-based models, all of which are based on public pre-trained models, using not only in-house but also public datasets.

\subsection{Datasets}
We used \textsc{mave}~\citep{mave}\footnote{\url{https://github.com/google-research-datasets/MAVE}} and our in-house product data for experiments. The \textsc{mave} dataset is designed to evaluate the extraction-based \textsc{pavi} models, while the in-house dataset is designed to evaluate classification-based models (Table~\ref{tab:dataset_stats}).

\begin{table*}[t]
\small
\tabcolsep3pt
    \centering
    \begin{tabular}{@{\,}p{0.17\linewidth}@{\quad}p{0.28\linewidth}@{\quad}l@{\quad}l@{\,}}
    \toprule
    \multicolumn{1}{c}{Title} & \multicolumn{1}{c}{Description} & \multicolumn{1}{c}{(original attribute-value info.)}& \multicolumn{1}{c}{Attribute-value pairs}\\\midrule

   Chicago Blackhawks Pet Dog Hockey Jersey LARGE &
 Chicago Blackhawks pet jersey - size LARGE. This great-looking jersey features screened-on logos on the sleeves and screened-on team name/number on the back.&
    \begin{tabular}[t]{@{}l@{}}

    $\langle$ Type, Jersey, 0, 34, 40 $\rangle$,\\
    $\langle$ Type, jersey, 1, 23, 29 $\rangle$,\\
    $\langle$ Type, jersey, 1, 63, 69 $\rangle$,\\
    $\langle$ Clothing Type, Jersey, 0, 34, 40 $\rangle$,\\
    $\langle$ Clothing Type, jersey, 1, 23, 29 $\rangle$,\\
    $\langle$ Clothing Type, jersey, 1, 63, 69 $\rangle$,\\
    $\langle$ Special use, None $\rangle$
    \end{tabular}
    &
    \begin{tabular}[t]{@{}l@{}}
    $\langle$ Type, Jersey $\rangle$,\\
    $\langle$ Type, jersey $\rangle$,\\
    $\langle$ Clothing Type, Jersey $\rangle$,\\
    $\langle$ Clothing Type, jersey $\rangle$,\\
    $\langle$ Special use, None $\rangle$
    \end{tabular}
    \\
    \midrule   
Northwave [northwave] Espresso Original Red Men's / Women's / Sneakers 25 - 27cm&
    Product description. These sneakers are the perfect accent for your feet and come in a soft red color. The sole is made of lightweight rubber to reduce weight. It is a popular color.&
    \begin{tabular}[t]{@{}l@{}}
    $\langle$ Shoe size (cm), 25.0 $\rangle$,\\
    $\langle$ Shoe size (cm), 26.0 $\rangle$,\\
    $\langle$ Shoe size (cm), 27.0 $\rangle$,\\
    $\langle$ Color, Red $\rangle$
    \end{tabular}
    &
    \begin{tabular}[t]{@{}l@{}}
    $\langle$ Shoe size (cm), 25.0 $\rangle$,\\
    $\langle$ Shoe size (cm), 26.0 $\rangle$,\\
    $\langle$ Shoe size (cm), 27.0 $\rangle$,\\
    $\langle$ Color, Red $\rangle$
    \end{tabular}
    \\\bottomrule
    \end{tabular}
    \caption{Example product data in \textsc{mave}  (top) and our in-house datasets (bottom, translated from Japanese), which include multi-attribute values (\textit{e.g., jersey}) and non-canonicalized values (\textit{e.g.}, 25).
    \textsc{mave} provides tuples of an attribute, value, paragraph ID, and the value's beginning and ending positions in the paragraph, while our in-house data provides canonicalized attribute values.
    To train \textsc{bert-ner}, 
    values in text are annotated by using the value positions in \textsc{mave} and by matching with the canonicalized values in the in-house data.
    To train \textsc{t5} and \textsc{bert-mlc} and evaluate all models, we use attribute-value pairs (right).}
    \label{tab:example_of_product_data}
\end{table*}

\smallskip\noindent\textbf{MAVE} dataset compiles the product data taken from Amazon Review Data~\citep{ni-etal-2019-justifying}. 
The dataset contains various kinds of products such as shoes, clothing, watches, books, and home decor decals.
Each example consists of 
product titles and descriptions, attribute, value, and span of the attribute value.
To construct such tuples, \citet{mave} trained five \textsc{aveqa} models~\citep{wang_2020} using a large amount of silver data where attribute values were annotated using manually tailored extraction rules. Then, they applied the trained models to the Amazon Review Data in order to detect spans of values corresponding to attributes given to the models.
To produce attribute value spans with high precision, they chose only attribute values that all five models extracted (positive). In addition, if no span is extracted from either model, and there is no extracted span from the extraction rules, they consider that there are no values for the attributes (negative); refer to  Table~\ref{tab:example_of_product_data} for example product data.
As a result, \textsc{mave} consists of 2,092,898 product data for training and 290,773 product data for testing.
Similar to~\citet{mave}, to make the training faster, we randomly selected 640,000 and 100,000 product data as the training and development sets from the original training data, respectively.
We used the test data in  \textsc{mave} for our evaluation as it is.
%
%

\smallskip\noindent\textbf{In-House Product Data} is taken from our e-commerce platform, Rakuten,\footnote{\url{https://www.rakuten.co.jp/}} which sells a wide range of products such as smartphones, car supplies, furniture, clothing, and kitchenware.
Each example consists of a tuple of title, description, and a set of attribute-value pairs.
The sellers assign products attribute-value pairs defined in the attribute taxonomy provided by the e-commerce platform. Since both attributes and values in the taxonomy are canonicalized, there 
exist
spelling gaps between values in the taxonomy and those in 
the product text
(\textit{e.g., Dolce \& Gabbana} in the taxonomy and \textit{D\&G} in the title). For experiments, among our in-house product data with one or more attribute-value pairs, we randomly sampled
640,000, 100,000, and 100,000 product data
for training, 
development,
and testing, respectively.

\begin{table*}[t]
    \small
    \centering
    \tabcolsep5.7pt
    \begin{tabular}{lrrrrrrr}\toprule
         & \multicolumn{3}{c}{\textsc{mave}} & \multicolumn{4}{c}{In-House Product Data} \\
        \cmidrule(l{2pt}r{2pt}){2-4}
        \cmidrule(l{2pt}r{2pt}){5-8}
         & \textsc{bert-ner} & \textsc{bert-mlc} & \textsc{t5} & \textsc{bert-ner} & \textsc{bert-mlc} & \textsc{bert-mlc} w/ \textsc{tax} & \textsc{t5}\\\midrule
    Training (10 epochs) &
    22 & 22 & 24 $\times$ 10 &
    22 & 22 & 22 & 24 $\times$ 10\\
    
    Inference (the dev set) &
    8 & 8 & 80 $\times$ 10 &
    8 & 8 & 8 & 80 $\times$ 10\\
    
    Inference (the test set)&
    1.6 & 1.6 & 16 $\times$ 6 &
    0.8 & 0.8 & 0.8 & 8 $\times$ 6\\
    \midrule
    Total &
    31.6 & 31.6 & 1,136 &
    30.8 & 30.8 & 30.8 & 1,088\\
    \bottomrule
    \end{tabular}
    \caption{GPU hours to perform our experiments. For \textsc{t5}, we finetune and evaluate *-first models for each composition (2 $\times$ 2 models). For the random ordering, we train three models for each composition (3 $\times$ 2 models), and check the performance on the development set to choose the model with the best micro F$_1$ for testing.}
    \label{tab:total_gpu_hours}
\end{table*}

\subsection{Models}

We compare the following models:

\smallskip\noindent\textbf{BERT-NER:} extraction-based model. On the top of \textsc{bert}, we place a classification layer that uses the outputs from the last layer of \textsc{bert} as feature representations of each subword. Each subword is classified into one of the labels. We employ \textsc{bilou} chunking scheme~\citep{sekine_1998,ratinov_2009}\yn{;} the total number of labels is $N \times 4 + 1$, where $N$ is the number of distinct attributes in the training data. We have used \textsc{bert} as the backbone here because the common extraction-based baseline~\citep{zheng_2018} uses classic BiLSTM-CRF as the backbone~\citep{Huang2015BidirectionalLM} and \textsc{bert}-based models outperform in \textsc{qa}-based models~\citep{wang_2020}; \textsc{bert-ner} can be a stronger and easily replicable baseline.

%


%
%
To annotate entities in text, we referred the beginning and ending positions in tuples for \textsc{mave}, and 
performed a dictionary matching for our in-house dataset. If annotations are overlapped, we keep the longest token length value, and drop all other overlapping values. For multi-attribute values, we adopt the most frequent attribute-value pair.

%
%

\smallskip\noindent\textbf{BERT-MLC:} classification-based model. We put a classification layer on the top of \textsc{bert}, and feed the embeddings of the \textsc{cls} token to the classification layer as a representation of given text~\citep{chen_2022}. The model predicts all possible attribute values from the representation through the classification layer. The total number of labels is the number of attribute values in the training data.

\smallskip\noindent\textbf{BERT-MLC w/ Tax:} the current state-of-the-art classification-based model that can be comparable with the other methods. 
We added to \textsc{bert-mlc} the label masking~\citep{chen_2022}, which leverages the skewed distributions of attributes in training and testing, using an attribute taxonomy defined for our in-house data. Although this is the state-of-the-art classification-based method, 
it \textbf{requires the attribute taxonomy as extra supervision}.
%
Since the \textsc{mave} dataset does not provide the attribute taxonomy, we train and evaluate this model only 
on our in-house dataset.

\smallskip\noindent\textbf{T5:} generation-based model of ours. We finetune \textsc{t5} on the training data 
obtained by each element in
$\lbrace$\textit{Attribute-then-value}, \textit{Value-then-attribute}$\rbrace\times\lbrace$\textit{Random}, \textit{Rare-first}, \textit{Common-first}$\rbrace$.
%
For random ordering, we create three training data with different 
random seeds, next train a model on each training data, and then chose the model that achieves the best micro F$_1$ on the development set.


\subsection{Implementations}

\begin{table*}[t]
\small
\tabcolsep4pt
    \centering
    \begin{tabular}{lllcccccccccccc}\toprule
    \multicolumn{3}{l}{} & \multicolumn{6}{c}{\textsc{mave}} & \multicolumn{6}{c}{In-House Product Data} \\
    \cmidrule(l{2pt}r{2pt}){4-9}
    \cmidrule(l{2pt}r{2pt}){10-15}
    \multicolumn{3}{l}{Models} & \multicolumn{3}{c}{Micro} & \multicolumn{3}{c}{Macro} &
    \multicolumn{3}{c}{Micro} & \multicolumn{3}{c}{Macro}\\
    \cmidrule(l{2pt}r{2pt}){4-6}
    \cmidrule(l{2pt}r{2pt}){7-9}
    \cmidrule(l{2pt}r{2pt}){10-12}
    \cmidrule(l{2pt}r{2pt}){13-15}
    \multicolumn{3}{l}{} & P (\%)& R (\%)& F$_1$ & P (\%) & R (\%) & F$_1$ & P (\%)& R (\%)& F$_1$ & P (\%)& R (\%)& F$_1$\\\midrule
    \multicolumn{3}{@{\,}l}{\textit{extraction-based}} \\
    \multicolumn{3}{l}{\textsc{bert-ner}} &
    \textbf{96.38} & 84.91 & 90.28 & 80.36 & 57.75 & 64.61 &
    \textbf{96.09} & 40.26 & 56.75 & 45.26 & 18.12 & 23.50\\
    \midrule


    \multicolumn{3}{@{\,}l}{\textit{classification-based}} \\
    \multicolumn{3}{l}{\textsc{bert-mlc}} &
    93.52 & 70.37 & 80.31 & 40.53 & 20.72 & 25.40 &
    \underline{94.53} & 74.43 & 83.29 & 40.81 & 18.05 & 22.82\\
    \multicolumn{3}{l}{\textsc{bert-mlc} w/ \textsc{tax}} &
    - & - & - & - & - & - &
    93.65 & 77.47 & 84.79 & 58.19 & 32.76 & 39.33\\
    \midrule


    \multicolumn{3}{@{\,}l}{\textit{generation-based (ours)}} \\
\textsc{t5} & $\langle$A, V$\rangle$ &
Rare-first &
\underline{95.45} & 91.70 & 93.54 & 77.57 & 64.35 & 68.97&
88.61 & 81.50 & \underline{84.91} & \textbf{66.33} & \textbf{47.25} & \textbf{53.10}\\

& & Common-first &
95.29 & \underline{92.16} & \underline{93.70} & 78.26 & 66.94 & 70.63&
85.30 & \underline{82.83} & 84.05 & 62.10 & 41.85 & 47.49\\

& & Random &
95.10 & 91.46 & 93.24 & 77.24 & 62.71 & 67.45&
87.73 & 81.41 & 84.45 & 61.64 & 42.47 & 47.92\\

& $\langle$V, A$\rangle$ &
Rare-first &
95.24 & 91.97 & 93.57 & \textbf{80.59} & \underline{68.02} & \underline{72.51} &
89.82 & 80.73 & \textbf{85.03} & \underline{65.73} & \underline{44.61} & \underline{50.93}\\
& & Common-first &
94.62 & \textbf{92.85} & \textbf{93.73} & 80.50 & \textbf{69.72} & \textbf{73.47} &
84.25 & \textbf{82.97} & 83.60 & 63.61 & 43.61 & 49.13\\
& & Random &
95.13 & 92.04 & 93.56 & \underline{80.56} & 67.28 & 71.83&
88.25 & 81.41 & 84.69 & 63.06 & 42.09 & 48.10\\

\bottomrule
    \end{tabular}
    \caption{Performance of each model on two \textsc{pavi} datasets; 
    The best score is in \textrm{bold face} and the second best score is underlined. \textsc{bert-mlc} w/ \textsc{tax} uses the extra supervision (taxonomy) for label masking, and it reduces the size of labels relevant to inputs from 14,829 to 405 on average.}
    \label{tab:old_performance}
\end{table*}

We implemented all models 
in PyTorch.\footnote{\url{https://pytorch.org/}}
We used \texttt{t5-base}\footnote{\label{t5_en}\url{https://huggingface.co/t5-base}} and \texttt{sonoisa/t5-base-japanese}\footnote{\label{t5_jp}\url{https://huggingface.co/sonoisa/t5-base-japanese}} in Transformers~\citep{wolf_2020}, both of which have 220M parameters,
as the pre-trained \textsc{t5} models for 
\textsc{mave} and 
our in-house data, respectively.
%
%
For training and testing, 
we used the default hyperparameters
provided with
each  model. We ran teacher forcing 
in training, and performed beam search of size four in testing.
%
For \textsc{bert}-based models,
we used \texttt{bert-base-cased}\footnote{\label{bert_en}\url{https://huggingface.co/bert-base-cased}} for \textsc{mave}, and \texttt{cl-tohoku/bert-base-japanese}\footnote{\label{bert_jp}\url{https://huggingface.co/cl-tohoku/bert-base-japanese}} for our in-house dataset,
both of which have 110M parameters.\footnote{
Training with BERT$_{\mathrm{large}}$ (330M parameters)
did not work for \textsc{bert-mlc} on either dataset; see
Table~\ref{tab:performance_bert_large} in Appendix.}
We set 0.1 of a dropout rate to a classification layer.

We use Adam~\citep{kingma2014adam} optimizer 
with learning rates shown in Table~\ref{tab:param_for_model_training} in Appendix.
We trained the models up to 10 epochs with a batch size of 32 and chose the models that perform the best micro F$_1$ on the development set.


\paragraph{Computing Infrastructure}

We used NVIDIA DGX A100 GPU on a Linux (Ubuntu) server with a AMD EPYC 7742 CPU at 2.25 GHz with 2 TB main memory for performing the experiments. Table~\ref{tab:total_gpu_hours} shows GPU hours taken for the experiments.

\subsection{Evaluation Measure}

Following the literature~\cite{xu_2019,wang_2020,mave,shinzato_2022,chen_2022}, we used micro and macro precision (P), recall (R), and F$_1$ 
as metrics. We compute macro performance in attribute-basis. Since the goal of \textsc{pavi} is not to detect spans of values in text but to assign attribute-value pairs to products, we \yn{pick one attribute-value pair from 
multiple identical attribute-value pairs}
in \textsc{mave}
(\textit{e.g.}, $\langle$Type, jersey$\rangle$ in Table~\ref{tab:example_of_product_data}). Note that we do not need this unification process for our in-house dataset because  it provides unique attribute-value pairs.

Since 
attribute values in the \textsc{mave} dataset are based on outputs from \textsc{qa}-based 
models~\citep{wang_2020} and those in our in-house data are assigned voluntarily by sellers on our marketplace, both datasets may contain some missing values.
To reduce the impact of those missing attribute-value pairs, we discard predicted attribute-value pairs if there are no ground truth labels for the attributes.


In the \textsc{mave} dataset, there are attributes whose values do not appear in the text (negative). For the ground truth with such no attribute values, models can predict no values (NN), or incorrect values (FP$_{\mathrm{n}}$) while for the ground truth with concrete attribute values, the model can predict no values (FN), correct values (TP), or incorrect values (FP$_{\mathrm{p}}$).
Based on those types of predicted values, P and R are computed as follows:
\begin{align*}
    & \text{P} = \frac{\lvert \text{TP} \rvert}{\lvert \text{TP} \rvert + \lvert \text{FP}_{\mathrm{p}} \rvert + \lvert \text{FP}_{\mathrm{n}} \rvert} \text{ , }
    \text{R} = \frac{\lvert \text{TP} \rvert}{\lvert \text{TP} \rvert + \lvert \text{FN} \rvert}.
\end{align*}
%
F$_1$ 
is computed as $2 \times \text{P} \times \text{R}$ / ($\text{P} + \text{R}$). Note that since there are no attributes with no values in our in-house dataset, the value of $\vert$FP$_{\mathrm{n}}\rvert$ is always 0.

\subsection{Results}

Table~\ref{tab:old_performance} shows the
performance of each model on \textsc{mave} and our in-house datasets.
Our generation-based models with *-first ordering mostly outperformed the extraction- and classification-based baselines in terms of F$_1$.\footnote{
The gap in performance may be partly attributed to the difference in the number of parameters in the base models. However, as shown in Table~\ref{tab:summary_of_requirements}, the generation model still has the advantage that it can address the challenges in the \textsc{pavi} task that the other approaches intrinsically cannot solve.
}
The differences between the \textbf{best} models and the baselines were significant ($p<0.0005$) under approximate randomized test~\citep{noreen}.
The higher recall of our generation-based models 
suggests the impact of capturing inter-value dependencies (\S~\ref{av_pair_ordering}).
%

The impact of the composition of attribute-value pairs depends on whether the output values are canonicalized. On the \textsc{mave} dataset, the models with the value-then-attribute composition outperformed 
those with attribute-then-value composition in terms of macro F$_1$. 
This is because all output values 
appear in the \textsc{mave} dataset. Thus, to the models, it is easier to generate values than attributes. Meanwhile, the advantage of value-then-attribute composition is smaller on our in-house dataset since there is no guarantee that the target values appear in the text as raw strings. 

The impact of the ordering of attribute-value pairs depends on the number of attribute-value pairs per example. On the in-house dataset, the models with rare-first ordering consistently outperformed those with common-first ordering in terms of F$_1$.
%
This result implies that decoding specific attribute-value pairs in advance is more helpful to generate general attribute-value pairs on the in-house dataset.
%
Meanwhile, there is no clear difference between the models with *-first orderings on the \textsc{mave} dataset, since the number of attribute-value pairs per example is small.

These results confirm that
%
the generative approach learns to flexibly perform canonicalization if it is required in the training data.\footnote{%
%
To make a more lenient comparison for \textsc{bert-ner} on the in-house dataset, 
we have also evaluated all models on attribute-value pairs in the test data whose attributes are observed in the training data of \textsc{bert-ner}.
On this test data, our generation-based model still outperformed the \textsc{bert-ner} and \textsc{bert-mlc} models; 
\textsc{t5} ($\langle$V, A$\rangle$, Rare-first) and \textsc{bert-ner} show the best micro (macro) F$_1$ of 85.75 (55.48) and 58.92 (30.17), respectively.}
Meanwhile, the performance of extraction- and classification-based approaches depends on whether the attribute-value pairs are canonicalized or not.

\begin{table}[t]
    \small
    \tabcolsep3pt
    \centering
    \begin{tabular}{cclccc}\toprule
    & \multicolumn{2}{c}{Models} & \multicolumn{3}{c}{\# of distinct values (med: 19)}\\
    & & & (19, $\infty$) & (0, 19] & all\\\midrule
   \# training & hi & \textsc{ner} & 90.5 / 80.1 & 90.2 / 69.3 & 90.5 / 77.3\\
     examples & & \textsc{mlc} & 80.7 / 40.2 & 85.5 / 34.9 & 80.8 / 38.9\\
     (med: 268)  & & \textsc{t5} & \textbf{93.9} / \textbf{86.9} & \textbf{94.4} / \textbf{78.2} & \textbf{93.9} / \textbf{84.7}\\
    \cmidrule{2-6}
    & lo & \textsc{ner} & 77.0 / 71.6 & 72.0 / 41.7 & 74.6 / 50.3\\
    & & \textsc{mlc} & 18.7 / \hfill 9.3 & 35.7 / 10.0 & 27.3 / \hfill 9.8\\
    & & \textsc{t5} & \textbf{81.1} / \textbf{76.7} & \textbf{79.4} / \textbf{54.8} & \textbf{80.3} / \textbf{61.1}\\
    \cmidrule{2-6}
    & all & \textsc{ner} & 90.4 / 78.0 & 87.0 / 49.9 &  90.3 / 64.6\\
    & & \textsc{mlc} & 80.4 / 32.6 & 78.4 / 17.4 & 80.3 / 25.4\\
    & & \textsc{t5} & \textbf{93.8} / \textbf{84.4} & \textbf{91.7} / \textbf{61.7} & \textbf{93.7} / \textbf{73.5}\\
\bottomrule
    \end{tabular}
    \caption{Micro / macro F$_1$ values of each approach on the \textsc{mave} dataset. ‘lo’ and ‘hi’ are intervals for the number of training examples, (0, 268] and (268, $\infty$], respectively. \textsc{t5} refers to the common-first model with $\langle$V, A$\rangle$ composition, which achieves the best micro F$_1$.}
    \label{tab:quantitative_comparision_for_mave}
\end{table}

\paragraph{Quantitative comparison of each approach}
To see the detailed behaviors of individual  approaches, we categorized the attributes in the \textsc{mave} and our in-house datasets
according to the number of training examples and the number of distinct values per attribute. We divide the attributes into four according to median frequency and number of values.

Tables~\ref{tab:quantitative_comparision_for_mave} and ~\ref{tab:quantitative_comparision} list micro and macro F$_1$ values of each approach for each category of attributes on the \textsc{mave} and our in-house datasets, respectively. From the table, we can see that \textsc{t5} shows the best performance in all categories. This suggests that \textsc{t5} is more robust than \textsc{bert-ner} and \textsc{bert-mlc} in the \textsc{pavi} task. We can also observe that the performance of \textsc{bert-mlc} drops significantly for attributes with a small number of training examples compared to those with a large number of training examples; the classification-based approach makes an effort to better classify more frequent attributes. Meanwhile, the performance drops of \textsc{bert-ner} and \textsc{t5} are more moderate than \textsc{bert-mlc}, especially on the \textsc{mave} dataset. Moreover, we can see that \textsc{t5} shows better micro F$_1$ for attributes that have a smaller number of distinct values on our in-house dataset, whereas 
it shows better micro F$_1$ for attributes that have a larger number of distinct values on the \textsc{mave} dataset.
This implies that, although it is easy for the generation-based approaches to extract diverse values from text, it is still difficult to canonicalize those diverse values.

\subsection{Analysis}

From the better macro F$_1$ of \textsc{t5} with *-first ordering than with random ordering, we confirmed that our generation-based models successfully
capture inter-value dependencies to decode attribute-value pairs. 
In what follows, we perform further analysis to see if the generative approach 
addresses the 
three challenges; namely,
unseen, multi-attribute (or nested), and canonicalized values  (Table~\ref{tab:summary_of_requirements}).

\paragraph{Can generative models identify unseen values?}
To see how effective our generative models are for unseen attribute values,
we 
compare its performance with \textsc{bert-ner} on attribute-value pairs in the test data that do not appear in the training data (13,578 and 491 unseen values exist in the \textsc{mave} and in-house datasets, respectively).

\begin{table}[t]
    \small
    \tabcolsep3.5pt
    \centering
    \begin{tabular}{cclccc}\toprule
    & \multicolumn{2}{c}{Models} & \multicolumn{3}{c}{\# of distinct values (med: 3)}\\
    & & & (3, $\infty$) & (0, 3] & all\\\midrule
    \# training & hi & \textsc{ner} & 56.4 / 29.6 & 62.9 / 31.6 & 56.8 / 30.2\\
    examples & & \textsc{mlc} & 83.5 / 34.1 & 82.1 / 44.6 & 83.4 / 37.2\\
    (med: 44) & & \textsc{t5} & \textbf{85.0} / \textbf{63.2} & \textbf{87.8} / \textbf{71.8} & \textbf{85.1} / \textbf{65.7}\\
    \cmidrule{2-6}
    & lo & \textsc{ner} & 25.2 / 14.8 & 27.7 / 14.0 & 26.8 / 14.3\\
    & & \textsc{mlc} & \hfill 5.0 / \hfill 1.6 & \hfill 8.9 / \hfill 3.1 & \hfill 7.4 / \hfill 2.7\\
    & & \textsc{t5} & \textbf{44.5} / \textbf{31.3} & \textbf{47.4} / \textbf{29.9} & \textbf{46.3} / \textbf{30.4}\\
    \cmidrule{2-6}
    & all & \textsc{ner} & 56.4 / 26.0 & 62.0 / 20.6 & 56.7 / 23.5\\
    & & \textsc{mlc} & 83.5 / 26.3 & 80.6 / 18.8 & 83.3 / 22.8\\
    & & \textsc{t5} & \textbf{84.9} / \textbf{55.5} & \textbf{86.8} / \textbf{45.7} & \textbf{85.0} / \textbf{50.9}\\
\bottomrule
    \end{tabular}
    \caption{Micro / macro F$_1$ values of each approach on the in-house dataset. ‘lo’ and ‘hi’ are intervals for the number of training examples, (0, 44] and (44, $\infty$], respectively. \textsc{t5} refers to the rare-first model with $\langle$V, A$\rangle$ composition, which achieves the best micro F$_1$.}
    \label{tab:quantitative_comparision}
\end{table}

\begin{table}[t]
\small
    \centering
    \tabcolsep3.1pt
    \begin{tabular}{lllcccc}\toprule
    \multicolumn{3}{l}{Models} & \multicolumn{2}{c}{\textsc{mave} F$_1$} & \multicolumn{2}{c}{In-House F$_1$} \\
    \cmidrule(l{2pt}r{2pt}){4-5}
    \cmidrule(l{2pt}r{2pt}){6-7}
    \multicolumn{3}{l}{} & Micro & Macro &
    Micro & Macro\\\midrule

\multicolumn{3}{l}{\textsc{bert-ner}} & 34.57 & 22.16 & 14.29 & 3.02\\\midrule
\textsc{t5} & $\langle$A, V$\rangle$ &
Rare-first & \textbf{38.21} & 27.87 & \textbf{19.44} & 5.08\\
& & Common-first & 37.34 & 29.02 & 17.03 & \textbf{6.55}\\
& & Random & 36.65 & 27.64 & 15.94 & 5.93\\
& $\langle$V, A$\rangle$ &
Rare-first & 37.44 & \underline{29.10} & 18.15 & 5.89\\
& & Common-first & \underline{38.19} & \textbf{31.22} & \underline{18.61} & \underline{6.15}\\
& & Random & 36.59 & 28.98 & 12.64 & 2.34

 \\ \bottomrule
    \end{tabular}
    \caption{Performance on 
    unseen values.
    The scores of \textsc{bert-mlc} models are 0.}
    \label{tab:unseen_performance}
\end{table}

Table~\ref{tab:unseen_performance} shows the results. 
We can see 
that the \textsc{t5} models outperform \textsc{bert-ner}, especially in terms of macro F$_1$.
Although the extraction-based approach can extract unseen values, the unified generative approach works better for extracting unseen values 
than the extraction-based approach.

\begin{table*}[t]
    \small
    \tabcolsep5pt
    \centering
    \begin{tabular}{lcc}\toprule
    Required processing    & Attribute-value pair & Text\\\midrule
    Understand structured values & $\langle$Series, iPhone (Apple)$\rangle$ & \textit{iPhone} 6S \textit{iPhone} Softbank...\\
    & $\langle$Chest (cm), 104 - 112$\rangle$ & ...Size [L] Chest \textit{110cm} Length 66cm... \\
    
    Refer to the world knowledge & $\langle$Sleeve length, Long$\rangle$ & Women's Trench \textit{Coat} Dark Brown...\\
    & $\langle$Indication, Rhinitis$\rangle$& For \textit{runny nose}, \textit{nasal congestion}, sore throat,...\\
    
    Recognize paraphrase & $\langle$Material, Polyurethane$\rangle$ & Material: \textit{PU} leather / Plastic\\
    & $\langle$Compatible brand, Galaxy S8 plus$\rangle$ & SC-03J \textit{Galaxy S8}$\mathit{+}$ Galaxy...\\
    
    Understand text & $\langle$Feature, With card holder$\rangle$ & \textit{The card slot is on the left}.\\
    & $\langle$With or without casters, With casters$\rangle$ & Table leg: Pipe, \textit{twin-wheel casters} with stopper...\\
    \bottomrule
    \end{tabular}
    \caption{Example of canonicalization that \textsc{t5} models need to perform to generate values that do not appear in text. Substrings in text that can be regarded as a clue to generate the values are in \textit{italic}.}
    \label{tab:example_of_required_canonicalization}
\end{table*}

\begin{table}[t]
\small
    \centering
    \tabcolsep3pt
    \begin{tabular}{lllcccc}\toprule
    \multicolumn{3}{l}{Models} & \multicolumn{2}{c}{\textsc{mave} F$_1$} & \multicolumn{2}{c}{In-House F$_1$} \\
    \cmidrule(l{2pt}r{2pt}){4-5}
    \cmidrule(l{2pt}r{2pt}){6-7}
    && & Micro & Macro &
    Micro & Macro \\\midrule

\multicolumn{3}{l}{\textsc{bert-ner}} & 47.85 & 35.60 & \textbf{81.35} & 45.54\\
\multicolumn{3}{l}{\textsc{bert-mlc}} & 68.79 & 24.95 & 76.43 & 30.47\\
\multicolumn{3}{l}{\textsc{bert-mlc} w/ \textsc{tax}} & - & - & 77.19 & 41.55 \\\midrule
\textsc{t5} & $\langle$A, V$\rangle$ &
Rare-first & 75.14 & 54.30 & 79.90 & \underline{58.44}\\
& & Common-first & 75.31 & 53.89 & 80.16 & 55.09\\
& & Random & 74.73 & 52.48 & \underline{80.45} & 53.50\\
& $\langle$V, A$\rangle$ &
Rare-first & \textbf{75.40} & \underline{56.11} & 80.13 & 57.81\\
& & Common-first & \underline{75.38} & \textbf{57.07} & 80.16 & \textbf{60.83}\\
& & Random & 74.97 & 54.28 & 80.18 & 56.08

 \\ \bottomrule
    \end{tabular}
\caption{Performance on attribute-value pairs that can be obtained only by identifying multi-attribute 
values.}
    \label{tab:performance_on_nested_values}
\end{table}

\paragraph{Can generative models identify multi-attribute values?}
Next, to see how effective our generative models are for identifying multi-attribute values, we compare its performance to the baselines
on attribute-value pairs in the test data that appear only as multi-attribute (or nested) values in input text. The number of such values in the \textsc{mave} and our in-house datasets is 60,832 and 15,843, respectively.

Table~\ref{tab:performance_on_nested_values} shows the results.
We can see that the \textsc{t5} models outperform all baselines in terms of macro F$_1$.
%
Although the classification-based models can identify multi-attribute values, the generative models outperformed
those models.


\paragraph{Can generative models identify canonicalized values?}
Lastly, to verify how effective our generative models are for identifying canonicalized values, we compare its performance with \textsc{bert-mlc} (w/ \textsc{tax}) on 207,997 attribute-value pairs whose values do not appear as raw strings in the corresponding product text in our in-house dataset. 

Table~\ref{tab:performance_on_canonicalized_values} shows the results. 
The \textsc{t5} models 
show comparable performance to and outperform the baselines in terms of micro 
 and macro F$_1$, respectively.
To see what types of canonicalization the \textsc{t5} models need to perform when the canonicalized values do not appear in the text, we manually inspect attribute-value pairs whose values do not appear in text on the development set.

Table~\ref{tab:example_of_required_canonicalization} exemplifies canonicalization that \textsc{t5} models need to perform. From the table, we can see that the canonicalization included understanding structure in values (labels) (\textit{e.g.,} \textit{iPhone} is a product of \textit{Apple}), referring the world knowledge (\textit{the coat} has \textit{long sleeves}), recognizing paraphrases (\textit{PU} is an abbreviation of \textit{polyurethane}), and understanding product descriptions (``\textit{the card slot is on the left}'' entails that the product \textit{has a card holder}).
%
%
We conclude that our generative model addressed all the challenges in the \textsc{pavi} task better than the other two approaches.

\begin{table}[t]
\small
    \centering
    \begin{tabular}{lllcc}\toprule
    \multicolumn{3}{l}{Models} & Micro F$_1$ & Macro F$_1$\\\midrule

\multicolumn{3}{l}{\textsc{bert-mlc}} & 72.49 & 20.10\\
\multicolumn{3}{l}{\textsc{bert-mlc} w/ \textsc{tax}} & \textbf{73.87} & 35.12\\\midrule
\textsc{t5} & $\langle$A, V$\rangle$ &
Rare-first & \underline{73.09} & \textbf{43.40}\\
& & Common-first & 71.91 & 39.07\\
& & Random & 72.48 & 39.19\\
 & $\langle$V, A$\rangle$ &
Rare-first & 72.93 & \underline{40.08}\\
& & Common-first & 71.20 & 37.95\\
& & Random & 72.30 & 37.20
 \\ \bottomrule
    \end{tabular}
\caption{Performance on attribute-value pairs whose values do not appear as raw strings in input text in our in-house test data. The score of \textsc{bert-ner} is 0.} 
    \label{tab:performance_on_canonicalized_values}
\end{table}

\section{Conclusions}
We have proposed a generative framework for product attribute-value identification (\textsc{pavi}), which is a task to return a set of attribute-value pairs from product text on e-commerce sites.
Our model can address the challenges of the \textsc{pavi} task; unseen values, multi-attribute values, and canonicalized values.
%
We finetune 
a pre-trained model \textsc{t5} 
to autoregressively decode a set of attribute-value pairs 
from the given product text.
To linearize the set of attribute-value pairs,
we explored two types of attribute-value composition and three types of the orderings of the attribute-value pairs.
%
Experimental results on two real-world datasets demonstrated that our generative approach 
outperformed the extraction- and classification-based baselines. 

We plan to augment the ability to decode unseen values by 
using a pluggable copy mechanism~\cite{liu-etal-2021-biocopy}.
We will evaluate our model
on another \textsc{pavi} setting where the target attribute(s) are given.

\section{Limitations}
Since our generative approach to product attribute-value identification autoregressively decodes a set of attribute-value pairs as a sequence, the inference is slow (Table~\ref{tab:total_gpu_hours}) and
how to linearize the set of attribute-value pairs in the training data will affect the performance (Table~\ref{tab:old_performance}).
The best way of composing an attribute-value pair and ordering the pairs will depend on the characteristics of the datasets such as the existence of canonicalized values and the number of  attribute-value pairs per example. 
Those who attempt to apply our method to their own datasets should keep this in mind.

\section*{Acknowledgements}
This work (second author) was partially supported by JSPS KAKENHI Grant Number 21H03494. We thank the anonymous reviewers for their hard work.

\bibliography{skeiji}
\bibliographystyle{acl_natbib}
\begin{table*}[t]
\small
\tabcolsep3.9pt
    \centering
    \begin{tabular}{lllcccccccccccc}\toprule
    \multicolumn{3}{l}{} & \multicolumn{6}{c}{\textsc{mave}} & \multicolumn{6}{c}{In-House Product Data} \\
    \cmidrule(l{2pt}r{2pt}){4-9}
    \cmidrule(l{2pt}r{2pt}){10-15}
    \multicolumn{3}{l}{Models} & \multicolumn{3}{c}{Micro} & \multicolumn{3}{c}{Macro} &
    \multicolumn{3}{c}{Micro} & \multicolumn{3}{c}{Macro}\\
    \cmidrule(l{2pt}r{2pt}){4-6}
    \cmidrule(l{2pt}r{2pt}){7-9}
    \cmidrule(l{2pt}r{2pt}){10-12}
    \cmidrule(l{2pt}r{2pt}){13-15}
    \multicolumn{3}{l}{} & P (\%)& R (\%)& F$_1$ & P (\%) & R (\%) & F$_1$ & P (\%)& R (\%)& F$_1$ & P (\%)& R (\%)& F$_1$\\\midrule

    \multicolumn{3}{@{\,}l}{\textit{extraction-based (large)}} \\
    \multicolumn{3}{l}{\textsc{bert-ner}} &
    \textbf{96.85} & 86.65 & 91.47 & 79.85 & 61.76 & 67.71 &
    \textbf{96.33} & 40.26 & 56.79 & 46.57 & 18.60 & 24.23\\
    \midrule


    \multicolumn{3}{@{\,}l}{\textit{classification-based (large)}} \\
    \multicolumn{3}{l}{\textsc{bert-mlc}} &
    93.19 & 51.63 & 66.45 & 17.31 &  7.40 & 9.49 &
    NaN & 0 & NaN & NaN & 0 & NaN\\
    \multicolumn{3}{l}{\textsc{bert-mlc} w/ \textsc{tax}} &
    - & - & - & - & - & - &
    \underline{94.36} & 77.99 & \textbf{85.40} & 52.36 &  28.01 & 33.92\\
    \midrule

    \multicolumn{3}{@{\,}l}{\textit{generation-based (ours)}} \\
\textsc{t5} & $\langle$A, V$\rangle$ &
Rare-first &
\underline{95.45} & 91.70 & 93.54 & 77.57 & 64.35 & 68.97&
88.61 & 81.50 & 84.91 & \textbf{66.33} & \textbf{47.25} & \textbf{53.10}\\

& & Common-first &
95.29 & \underline{92.16} & \underline{93.70} & 78.26 & 66.94 & 70.63&
85.30 & \underline{82.83} & 84.05 & 62.10 & 41.85 & 47.49\\

& & Random &
95.10 & 91.46 & 93.24 & 77.24 & 62.71 & 67.45&
87.73 & 81.41 & 84.45 & 61.64 & 42.47 & 47.92\\

& $\langle$V, A$\rangle$ &
Rare-first &
95.24 & 91.97 & 93.57 & \textbf{80.59} & \underline{68.02} & \underline{72.51} &
89.82 & 80.73 & \underline{85.03} & \underline{65.73} & \underline{44.61} & \underline{50.93}\\
& & Common-first &
94.62 & \textbf{92.85} & \textbf{93.73} & 80.50 & \textbf{69.72} & \textbf{73.47} &
84.25 & \textbf{82.97} & 83.60 & 63.61 & 43.61 & 49.13\\
& & Random &
95.13 & 92.04 & 93.56 & \underline{80.56} & 67.28 & 71.83&
88.25 & 81.41 & 84.69 & 63.06 & 42.09 & 48.10\\

\bottomrule
    \end{tabular}
    \caption{Performance of each model on two  \textsc{pavi} datasets. We used \textsc{bert}$_{\mathrm{large}}$ as a base model for extraction- and classification-based approaches.
    The best score is in \textrm{bold face} and the second best score is underlined. \textsc{bert-mlc} w/ \textsc{tax} uses the extra supervision (taxonomy) for label masking, and it reduces the size of labels relevant to inputs from 14,829 to 405 on average.}
    \label{tab:performance_bert_large}
\end{table*}

\begin{table}[t]
\small
    \centering
    \tabcolsep2pt
    \begin{tabular}{lccc}\toprule
Hyperparameters                  & \textsc{bert-ner} & \textsc{bert-mlc} & \textsc{t5}\\\midrule
Max token length (encoder) & 512 & 512 & 512 \\
Max token length (decoder) & n/a & n/a& 256 \\
Epoch                      & 10 & 10 & 10 \\
Batch size                 & 32 & 32 & 32 \\
Dropout rate (classifier)  & 0.1 & 0.1 & n/a \\
Learning rate              & 5e-5 & 5e-5 & 3e-4 \\
Weight decay               & 0 & 0 & 0 \\
\bottomrule
    \end{tabular}
    \caption{Hyperparameters for training models.}
\label{tab:param_for_model_training}
\end{table}

\appendix

\section{Final Hyperparameters Used for Each Model}
Table~\ref{tab:param_for_model_training} shows the hyperparameters we used for training models. Other than those, we follow the default hyperparameters of \textsc{t5}\footnoteref{t5_en}~\footnoteref{t5_jp} and \textsc{bert}\footnoteref{bert_en}~\footnoteref{bert_jp} available from the HuggingFace models.

\section{Performance of Models Using \textsc{bert}$_{\mathrm{large}}$}

Table~\ref{tab:performance_bert_large} shows the performance of models when we use \textsc{bert}$_{\mathrm{large}}$
as the base model for extraction- and classification-based approaches.
We adopt \texttt{bert-large-cased}\footnote{\url{https://huggingface.co/bert-large-cased}} for \textsc{mave} and \texttt{cl-tohoku/bert-large-japanese}\footnote{\url{https://huggingface.co/cl-tohoku/bert-large-japanese}} for our in-house data.
From the table, we can see that training \textsc{bert-mlc} did not work well on both datasets. Especially, we cannot compute the performance on our in-house data because the model did not predict any attribute-value pairs for all inputs.
Although \textsc{bert}$_{\mathrm{large}}$ has a larger number of parameters (330M) than the \textsc{t5} models (220M), \textsc{bert-ner} based on \textsc{bert}$_{\mathrm{large}}$ still shows lower performance than our generative models on both datasets. This result means that our generative approach is more effective in the \textsc{pavi} task than the extraction-based approaches based on \textsc{bert-ner}.
Meanwhile, \textsc{bert-mlc} w/ \textsc{tax} shows a slightly better micro F$_1$ score than ours. Given that it requires an attribute taxonomy as the extra supervision and exhibits low macro F$_1$, the generative approach is sufficiently comparable to the classification-based approach.

\end{document}